\documentclass{article}
\usepackage[utf8]{inputenc}
\usepackage{graphicx} 
\usepackage{subfigure} 

\usepackage{natbib}

\usepackage{algorithm}
\usepackage{algorithmic}
\usepackage{hyperref}

\usepackage[accepted]{icml2013} 

\usepackage{fancyvrb}
\usepackage{amsmath}
\icmltitlerunning{Text segmentation with character-level text embeddings}

\begin{document} 

\twocolumn[
\icmltitle{Text segmentation with character-level text embeddings}

\icmlauthor{Grzegorz Chrupała}{g.chrupala@uvt.nl}
\icmladdress{Tilburg Center for Cognition and Communication,
            Tilburg University,
            5000 LE Tilburg,
            The Netherlands}

\icmlkeywords{character-level, simple recurrent network, sequence labeling}

\vskip 0.3in
]

\begin{abstract}
  Learning word representations has recently seen much success in
  computational linguistics.  However, assuming sequences of word
  tokens as input to linguistic analysis is often unjustified. For
  many languages word segmentation is a non-trivial task and naturally
  occurring text is sometimes a mixture of natural language strings and
  other character data.  We propose to learn text representations
  directly from raw character sequences by training a Simple Recurrent
  Network to predict the next character in text. The network uses its
  hidden layer to evolve abstract representations of the character
  sequences it sees.  To demonstrate the usefulness of the learned
  text embeddings, we use them as features in a supervised character
  level text segmentation and labeling task: recognizing spans of text
  containing programming language code. By using the embeddings as
  features we are able to substantially improve over a baseline which
  uses only surface character n-grams.
\end{abstract}

\section{Introduction}
\label{sec:introduction}
The majority of representations of text used in computational
linguistics are based on words as the smallest units. Automatically
induced word representations such as distributional word classes or
distributed low-dimensional word embeddings have recently seen much
attention and have been successfully used to provide generalization
over surface word forms
\citep{collobert:2008,turian2010word,chrupala2011efficient,collobert:2011b,socher2012semantic,
  chen2013expressive}.

In some cases, however, words may be not the most appropriate atomic
unit to assume as input to linguistic analysis. In polysynthetic and
agglutinative languages orthographic words are typically too large as
a basic unit, as they often correspond to whole English phrases or
sentences. Even in languages where the word is approximately the right
level of granularity we often encounter text which is a mixture of
natural language strings and other character data. One example is the
type of text used for the experiments in this paper: posts on a
question-answering forum, written in English, with segments of
programming language example code embedded within the text.

In order to address this issue we propose to induce text
representations directly from raw character strings. This sidesteps
the issue of what counts as a word and whether orthographic words are the
right level of granularity. At the same time, we can elegantly deal
with character data which contains a mixture of languages, or domains,
with differing characteristics. In our particular data, it would not
be feasible to use words as the basic units. In order to split text
from this domain into words, we first need to segment it into
fragments consisting of natural language versus fragments consisting
of programming code snippets, since different tokenization rules apply
to each type of segment.

Our representations correspond to the activation of the hidden layer
in a simple recurrent neural network (SRN) \citep{elman1990finding,elman1991distributed}. The
network is sequentially presented with raw text and
learns to predict the next character in the sequence. It uses the
units in the hidden layer to store a generalized representation of the
recent history. After training the network on large amounts on
unlabeled text, we can run it on unseen character sequences, record
the activation of the hidden layer and use it as a representation
which generalizes over text strings.

We test these representations on a character-level sequence labeling
task. We collected a large number of posts to a programming
question-answering forum which consist of English text with embedded
code samples. Most of these code segments are delimited with HTML
tags and we use this markup to derive labels for supervised learning. 
As a baseline we train a Conditional Random Field model with character n-gram features
We then compare to it the same baseline model enriched with features derived
from the learned SRN text representations. We show that the generalization
provided by the additional features substantially improves 
performance: adding these features has similar effect to quadrupling
the amount of training data given to the baseline model.

\section{Simple Recurrent Networks}

Text-representations based on recurrent networks will be discussed in
full detail elsewhere. Here we provide a compact overview of the
aspects most relevant to the text segmentation task.

Simple recurrent neural networks (SRNs) were first introduced by
\citet{elman1990finding,elman1991distributed}. The units in the hidden layer
at time $t$ receive incoming connections from the input units at time
$t$ and also from the hidden units at the previous time step
$t-1$. The hidden layer then predicts the state of the output units at
the next time step $t+1$. The weights at each time step are
shared. The recurrent connections endow the network with memory which
allows it to store a representation of the history of the inputs
received in the past.

We denote the input layer as $w$, the hidden layer as $s$ and the
output layer as $y$. All these layers are indexed by the time
parameter $t$: the input vector to the network at time $t$ is $w(t)$,
the state of the hidden layer is $s(t)$ and the output vector is
$y(t)$. 

The input vector $w(t)$ represents
the input element at current time step, in our case the current
character.  The output vector $y(t)$ represents the predicted
probabilities for the next character in the sequence.

The activation of a hidden unit is a function of the current input
and the state of the hidden layer at the previous time step:
 $t-1$:
\begin{equation}
  s_j(t) = f\left(\sum_{i=1}^I w_i(t)U_{ji} + \sum_{l=1}^J s_j(t-1)W_{jl}\right)
\end{equation}
where $f$ is the sigmoid function:
\begin{equation}
  f(a) = \frac{1}{1+\exp(-a)},
\end{equation}
and $U_{ji}$ is the weight between input component $i$ and hidden unit
$j$, while $W_{jl}$ is the weight between hidden unit $l$ and
hidden unit $j$.

The components of the output vector are defined as:
\begin{equation}
  y_k(t) = g\left(\sum_{j=1}^J s_j(t) V_{kj}\right),
\end{equation}
where $g$ is the softmax function over the output components:
\begin{equation}
  g(z) = \frac{\exp(z)}{\sum_{z'} \exp(z')},
\end{equation}
and $V_{kj}$ is the weight between hidden unit $j$ and output
unit $k$.

SRN weights can be trained using backpropagation through time (BPTT)
\cite{rumelhart1986learning}. With BPTT a recurrent network with $n$
time steps is treated as a feedforward network with $n$ hidden layers
with weights shared at each level, and trained with standard
backpropagation.

BPTT is known to be prone to problems with exploding or vanishing
gradients. However, as shown by \citep{mikolov2010recurrent}, 
for time-dependencies of moderate length they are
competitive when applied to language modeling.  Word-level SRN
language models are state of the art, especially when used in
combination with n-grams. 

Our interest here, however, lies not
so much in using SRNs for language modeling per se, but rather in
exploiting the representation that the SRN develops while learning to
model language. Since it does not have the capacity to store explicit
history statistics like an n-gram model, it is forced to generalize
over histories. As we shall see, the ability to create such
generalizations has uses which go beyond predicting the next character
in a string.



\section{Recognizing and labeling code segments}
We argued in the Section~\ref{sec:introduction} that there are often cases where using
words as the minimum units of analysis is undesirable or
inapplicable. Here we focus on one such scenario. Documents such as
emails in a software development team or bug reports in an issue
tracker are typically mostly written in a natural language (e.g.\ English)  
but have also embedded within them fragments of programming source
code, as well as other miscellaneous non-linguistic
character data such as error messages, stack traces or program
output. Frequently these documents are stored in a plain text format,
and the boundaries between these different text segment, while evident
to a human, are not explicitly indicated. When processing such
documents it would be useful to be able to preprocess them and
recognize and label non-linguistic segments as such. 
We develop such a labeler by training it on a large set of documents
where segments are explicitly marked up.

We collected questions posted to
\href{http://stackoverflow.com}{Stackoverflow.com} between February and Jun
2011.  Stackoverflow is not a pure forum: it also incorporates
features of a wiki, such that posted questions can be edited by other
users and formating, clarity and other issues can be improved. This
results in a dataset where code blocks are quite reliably marked as
such via HTML tags. Short inline code fragments are also often marked up but much
less reliably.

\begin{figure}
\centering
  \includegraphics[width=\columnwidth]{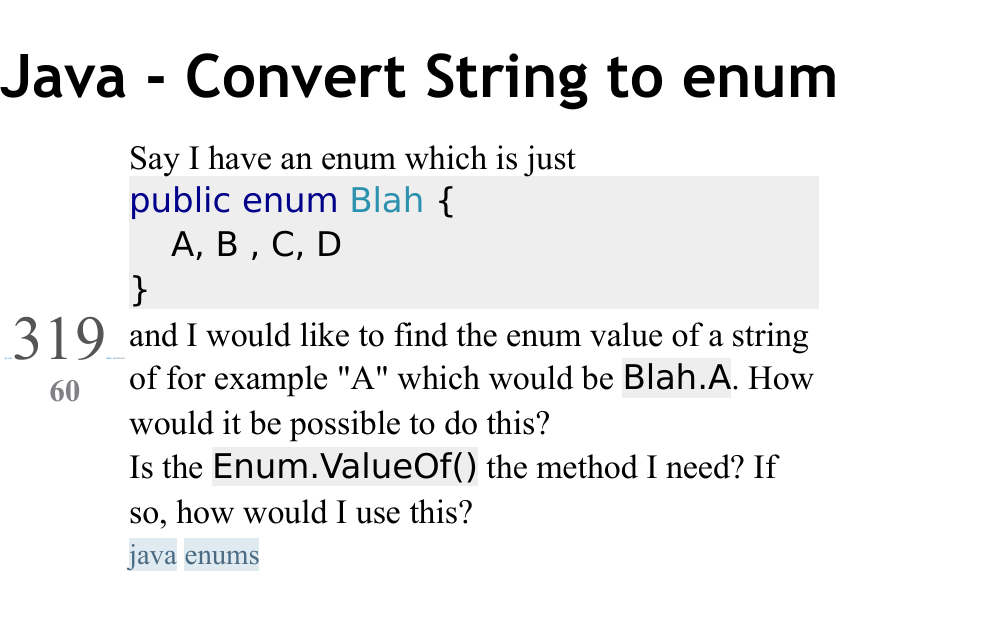}
  \vskip -0.5cm
  \caption{Example Stackoverflow post.}
  \label{fig:so-post}
\end{figure}

Figure~\ref{fig:so-post} shows an example post to the Stackoverflow
forum: it has one block code segment and two inline segments. 

We convert the marked-up text into labeled character sequences using
the BIO scheme, commonly used in NLP sequence labeling tasks.
We distinguishing between block and inline code
segments. Labels starting with {\tt B-} indicate the beginning of a
segment, while the ones staring with {\tt I-} stand for continuation of
segments.  Figure \ref{fig:bio} shows an example.
\begin{figure}
  \begin{center}
    \begin{tt}
      \begin{tabular}{|ll|}\hline
        j & O \\
        u & O \\
        s & O \\
        t & O \\
        ¶ & O \\
        p & B-BLOCK \\
        u & I-BLOCK \\
        b & I-BLOCK \\
        l & I-BLOCK \\ 
        i & I-BLOCK \\
        c  & I-BLOCK\\
        \textvisiblespace & I-BLOCK \\\hline
      \end{tabular}
      \begin{tabular}{|ll|}\hline
        b & O \\
        e & O \\
        \textvisiblespace & O \\
        B & B-INLINE \\
        l & I-INLINE \\
        a & I-INLINE \\
        h & I-INLINE \\
        . & I-INLINE \\
        A & I-INLINE\\
        . & O\\
        H & O \\
        o & O \\\hline
      \end{tabular}
    \end{tt}
    \caption{Example sequence labeling derived from the example
      Stackoverflow post.}
    \label{fig:bio}
  \end{center}
\end{figure}
Using such labeled data we can create a basic labeler by training a
standard linear chain Conditional Random Field (we use the Wapiti
implementation of \citet{lavergne2010practical}).  Our main interest here
lies in determining how much text representations learned by SRNs from
unlabeled data can help the performance of such a labeler.

\section{Experimental evaluation}

We create the following disjoint subsets of data from the Stackoverflow
collection:
\begin{itemize}
\item 465 million characters unlabeled data set for learning text
  representations (called {\sc large})\footnote{We do have the
    automatically derived labels for this dataset. We did not
    use them for two reasons: (i) the prohibitive RAM
    requirements for a CRF with such a large amount of training
    examples; (ii) more importantly, we were interested in the much
    more common scenario where only a limited amount of labeled data
    is available. }
\item 10 million characters training set. We use this data (with labels) to train
  the CRF model. We use it also (without the labels) to learn an alternative model of text
  representations (called {\sc small}) 
\item 2 million characters labeled development set for tuning the CRF model
\item 2 million characters labeled test set for the final evaluation
\end{itemize}

\subsection{Training SRNs}
As our SRN implementation we use a customized version of
\citet{mikolov2010recurrent}'s RNNLM toolkit.  We trained two separate
SRN models, {\sc large} on the full 465 million-character data set and
{\sc small} on the 10-million-character data set. The segmentation
labels are not used for SRN training.
Input character are represented as one-hot vectors. For both models we
use 400 hidden units, and 10 steps of BPTT.  We trained the {\sc
  large} model for 6 iterations (this took almost 2 CPU-months). The
{\sc small} model was trained until convergence, which took 13
iterations (less than a day).

In order to understand better the nature of the learned text
embeddings we performed the following analysis: After training the {\sc large}
SRN model we run it in prediction mode 
on the initial portion of the development data. We record
the activation of the hidden layer at each position in the text as the
network is making a prediction at this position. We then sample 
positions 100 characters apart, and for each of them find the
four nearest neighbors in the initial 10000 characters of the
development data. We use cosine of the angle between the hidden layer
activation vectors as a similarity
metric. 
\begin{figure}[!t]
\begin{small}
\begin{Verbatim}[frame=single, framesep=5mm]
  esetMetaData();¶    };¶¶    func
  gFucntions();  ¶    };¶¶    func
  dlerToTabs();¶    };  ¶¶    func
   }           ¶    };  ¶¶    func
          metaConstruct;¶¶    func

  n-laptop": {"last_share": 130738
  ierre-pc": {"last_share": 130744
  d-laptop": {"last_share": 130744
  laptop": {"last_share": 13074434
  erre-pc": {"last_share": 1307441

   data table has integer values a
  ,2,3,4,5. For all these values I
  ere i can add more connections s
  eating lots of private methods a
  or more different data sources c

  e given URL.I'd like to change t
  e = SqlPersist¶¶¶When I remove t
  sources explaining how to save f
  basic knowledge doesn't enable m
  eDirectory, but I need to save t
\end{Verbatim}
\end{small}
  \caption{Examples of nearest neighbors as measured by cosine between
  hidden layer activation vectors.}
  \label{fig:nearest-neighbors}
\end{figure}
Figure~\ref{fig:nearest-neighbors} shows four examples from the
qualitative analysis of this data: the
first row in each example is the sampled position, the next four rows
are its nearest neighbors. The activation was recorded as the network
was predicting the last character in each row.

We often find that the nearest neighbors simply share the literal
history: e.g.\ {\tt ¶¶    func} in the first example. However, the network
is also capable of generalizing over the surface form,
as can be seen in the second example, where the numerals in the string
suffix vary. 
The last two examples show an even higher level of generalization. Here the
surface forms of the strings are different, but they are related
semantically: they all end with plural nouns and with transitive verbs
respectively.


\subsection{Features}

\paragraph{Baseline feature set}
We used simple character n-gram features centered around the focus
character for our baseline labeler. Table~\ref{tab:features-baseline}
shows an example. 
\begin{table}
  \begin{center}
    \caption{Features extracted from the character sequence
      {\tt just¶public} while focused on the character
      '{\tt\bf{p}}'. }
    \label{tab:features-baseline}
    \vskip 0.3cm
    \begin{tabular}{l|c}
      Unigram  & \tt t ¶ p u b\\
      Bigram   & \tt ¶p pu     \\
      Trigram  & \tt ¶pu      \\
      Fourgram & \tt t¶pu ¶pub \\
      Fivegram & \tt t¶pub
    \end{tabular}
  \end{center}
\end{table}

\paragraph{Augmented feature set} 
The second feature set is the baseline feature set augmented with
features derived from the text representations learned by the SRN.  The
representation corresponds to the activation of the hidden layer.
After training the network, we freeze the weights, and run it in prediction model
on the training and development/test data, and record the activation of the hidden
layer at each position in the string as the network tries to predict the next
character. 

We convert the activation vector to the binary indicator features
required by our CRF toolkit as follows: for each of the $K=10$ most active units out of total
$J=400$ hidden units, we create features $(f(1) \ldots f(K))$ defined
as: 
\begin{equation}
f(k) = 
\begin{cases}
  1 & \mbox{ if } s_{j(k)} > 0.5 \\
  0 & \mbox{ otherwise }
\end{cases}
\end{equation}
where $j(k)$ returns the index of the $k$\textsuperscript{th} most
active unit.
We set $K=10$ based on preliminary experiments which indicated that
increasing this value has little effect on performance. This is due to
the fact that in the network only few hidden units are typically
active, with the large majority of activations close to zero. 

\subsection{Results}
Table~\ref{tab:baseline-dev} shows the results on the development set
for the model trained with baseline features. The F1 score is computed
segment-wise: any mistake in detecting segment boundaries correctly
results in a penalty.

\begin{table}
  \centering
  \caption{Results on the development set with baseline features.}
  \vskip 0.2cm
  \begin{tabular}{l|r|r|r}
  Label      & \% Precision & \% Recall & \% F1    \\\hline
  \sc block  & 88.96     & 87.91  & 88.43 \\
  \sc inline & 35.87     &  8.88  & 14.23 \\\hline
  Overall    & 81.54     & 56.80  & 66.96 \\
\end{tabular}  
  \label{tab:baseline-dev}
\end{table}

While the performance for
block segments is reasonable, for inline segments it is very low:
especially as measured by recall. Inspecting the data we determined that
inline segment marking in Stackoverflow posts is very inconsistent. 
A proper evaluation of performance on inline segments would thus
require very labor intensive manual correction of this type of label
in our dataset.  
We thus focus mostly on the performance with the much more reliable block
segment labeling in the remainder of the paper.

In order to get a picture of the influence on the performance of the
number of both labeled and unlabeled examples, we trained the CRF model
while repeatedly doubling the amount of labeled training data: we
start with 12.5\% of the full 10 million characters, and continue with
25\%, 50\% and 100\%. We used three feature sets: the baseline
features set, as well as two augmented feature sets, {\sc small} and
{\sc large}, corresponding to the SRN being trained on 10 million and
465 million characters respectively.
 
\begin{figure}
  \centering
  \includegraphics[width=\columnwidth]{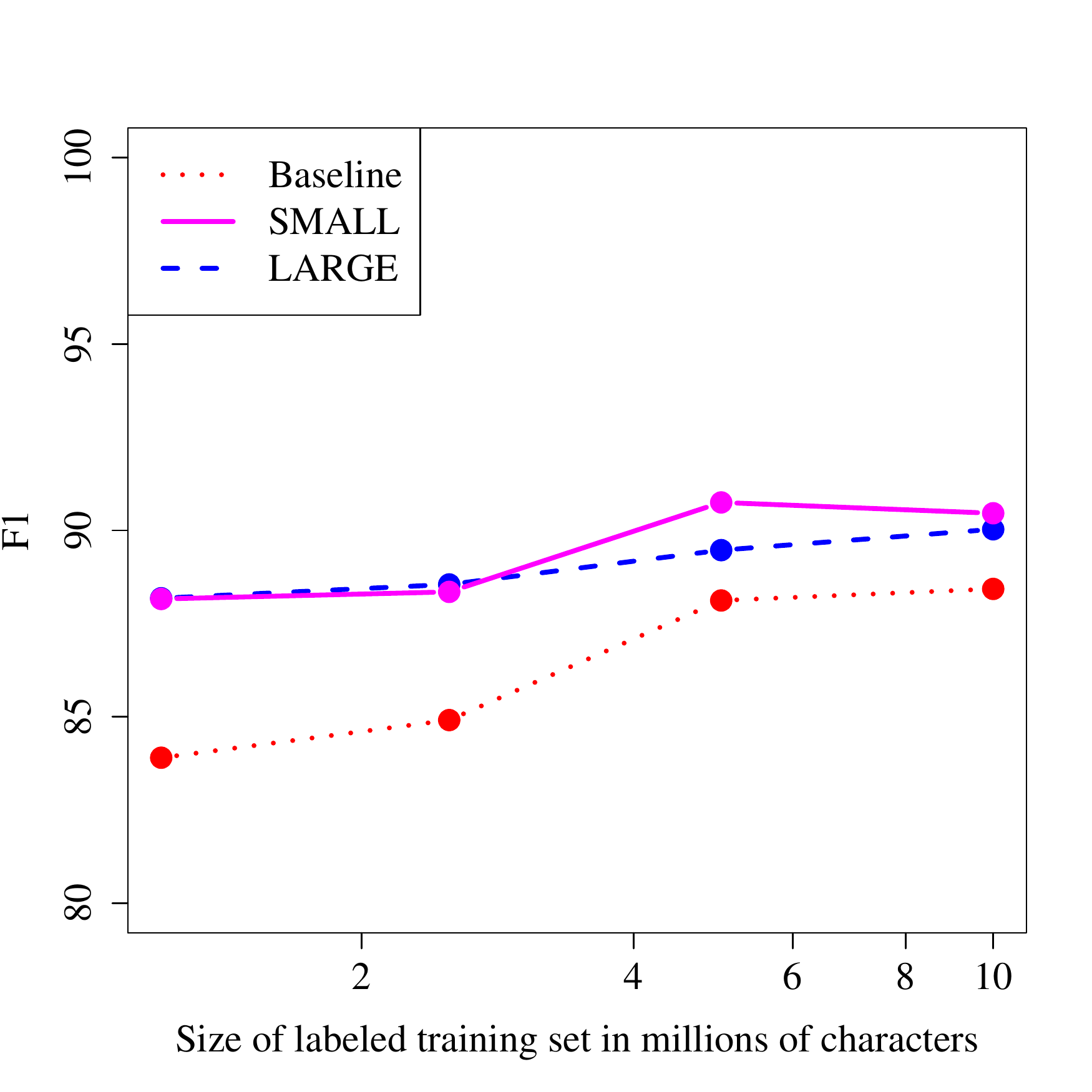}
  \caption{Block segment performance of baseline and augmented feature
    sets on the development set.}
  \label{fig:ablations}
\end{figure}

Figure~\ref{fig:ablations} shows the performance on the
development set of the labeler with each of these training sets and
feature sets. 

As can be appreciated from the plot, both the augmented feature sets
boost performance to the degree roughly corresponding to quadrupling
the amount of labeled training examples: i.e.\ using the augmented
feature set with 12.5\% of the labeled training examples results in
an F1 score approximately the same as using the baseline feature set
with 50\% of the labeled examples.

It is also interesting that the extra features coming from the {\sc
  small} SRN model do no worse than the ones from the {\sc large}
model. This seems to indicate the the performance boost from the
SRN features are largely exclusively due to these features being more
expressive and not due to the extra unlabeled text that they were
derived from. 

On one hand this is good news as it means that we can gain a large
boost in performance by training an SRN on a moderate amount of data,
and spending CPU-months on processing huge datasets is not necessary.

On the other hand, however, we would like to get an additional improvement
from large datasets whenever we \emph{can} afford the additional CPU
time. We were not able to show this benefit for the data in this
study. Also in terms of the SRN language model quality as evaluated on the
development dataset, the much larger amount of data did not show a
substantial benefit: model perplexity was 4.24 with the {\sc small}
model and 4.11 with the {\sc large} model. This may be related to
temporal concept drift across the Stackoverflow posts causing a
divergence between the large dataset and the development and
test datasets. Clearly these issues deserve to be examined more
exhaustively in future.

Table~\ref{tab:test-full} shows the performance on block-level
segmentation on the final test set when using the three feature
sets with 100\% of the labeled training set. The picture is similar to what 
we saw when analyzing results on development data. Here the
SRN features from {\sc large} unlabeled data outperform
the SRN features from small data only slightly.
Appendix~\ref{appendix:a} contains a more complete set of evaluation
results.

\begin{table}
  \centering
  \caption{Block segment performance on final test set with three
    feature sets. }
\vskip 0.2cm
  \begin{tabular}{l|r|r|r}
    Model     & \% Precision & \% Recall & \% F1 \\\hline
    Baseline  & 85.62        & 87.29     & 86.45 \\
    {\sc Small}     & 90.28        & 90.42     & 90.35 \\
    {\sc Large}     & 90.75        & 91.15     & 90.95 \\
  \end{tabular}
  \label{tab:test-full}
\end{table}


\section{Related work}

There is a growing body of research on using \emph{word} embeddings as
features in NLP tasks. \citet{collobert:2008} and \citet{collobert:2011b} use them
in a setting where a number of levels of linguistic annotation are
learned jointly: part-of-speech tagging, chunking, named-entity
labeling and semantic role labeling. \citet{collobert2011deep} applies
the same technique to discriminative parsing. \citet{turian2010word}
test a number of word representations including embeddings produced by
neural language models on syntactic chunking and named entity
recognition.  \citet{SocherEtAl2011:PoolRAE,socher2012semantic}
recursively compose word embeddings to produce distributed
representations of phrases: these in turn are tested on a number of
tasks such as prediction of phrase sentiment polarity or paraphrase
detection. Finally, \citet{chen2013expressive} compare a number of
word embedding types on a battery of NLP word classifications tasks.

We are not aware of any work on character-level word
embeddings. \citet{mikolov2012subword} investigate subword level SRNs
as language models, but do not discuss the character of the learned
text representations.

We also do now know of any work on learning to detect and
label code segments in raw text. However,
\cite{bettenburg2008extracting} describe a system called \mbox{\em infoZilla}
which uses hand-written rules to extract source code fragments, stack
traces, patches and enumerations from bug reports. In contrast, here we
leverage the Stackoverflow dataset to learn how to perform
a similar task automatically. 

\section{Conclusion}

In this study we created datasets and models for the task of
supervised learning to detect and label code blocks in raw
text.  Another major contribution of our
research is to provide evidence that character-level text embeddings
are useful representations for segmentation and labeling of raw text
data. We also have preliminary indications that these representations
are applicable in other similar tasks.

In this paper we have only scratched the surface and there are many
important issues that we are planning to investigate in future work. 
Firstly, a version of recurrent networks with multiplicative connections was introduced by
\citet{sutskever2011generating} and trained on the text of
Wikipedia. We would like to see how embeddings from that model perform.

Secondly, in the current paper we adopted a strictly modular
setup, where text representations are trained purely on the character
prediction task, and then used as features in a separate supervised
classification step. This approach has the merit that the same text
embeddings can be reused for multiple
tasks. Nevertheless it would also be interesting to investigate the
behavior of a joint model, which learns to predict characters and
their labels simultaneously.


\bibliographystyle{icml2013}
\bibliography{biblio}

\appendix
\section{Appendix}
\label{appendix:a}

\begin{table}[!h]
  \centering
  \caption{Evaluation results on the test set with full (10 million
    characters) training set and baseline featurset. }
\vskip 0.5cm
  \begin{tabular}{lrrr}
        & \% Precision &  \% Recall  & \% F1 \\\hline
BLOCK   &   85.62 &  87.29 &  86.45 \\
INLINE  &   36.60 &  10.24 &  16.00 \\\hline
Overall &   78.22 &  57.01 &  65.95
  \end{tabular}
  \label{tab:baseline-full}
\end{table}

\begin{table}[!h]
  \centering
  \caption{Evaluation results on the test set with full (10 million
    characters) training set and {\sc small} featurset. }
\vskip 0.5cm
  \begin{tabular}{lrrr}
        & \% Precision &  \% Recall  & \% F1 \\\hline
BLOCK   &   90.28 &  90.42 &  90.35 \\
INLINE  &   34.95 &  11.34 &  17.12 \\\hline
Overall &   80.69 &  59.34 &  68.38
\end{tabular}
  \label{tab:small-full}
\end{table}

\begin{table}[!h]
  \centering
  \caption{Evaluation results on the test set with full (10 million
    characters) training set and {\sc large} featurset. }
\vskip 0.5cm
  \begin{tabular}{lrrr}
        & \% Precision &  \% Recall  & \% F1 \\\hline
BLOCK   &   90.75 &  91.15 &  90.95 \\
INLINE  &   35.62 &  11.58 &  17.47 \\\hline
Overall &   81.20 &  59.87 &  68.92
\end{tabular}
  \label{tab:large-full}
\end{table}

\begin{figure}[!h]
  \centering
  \includegraphics[width=\columnwidth]{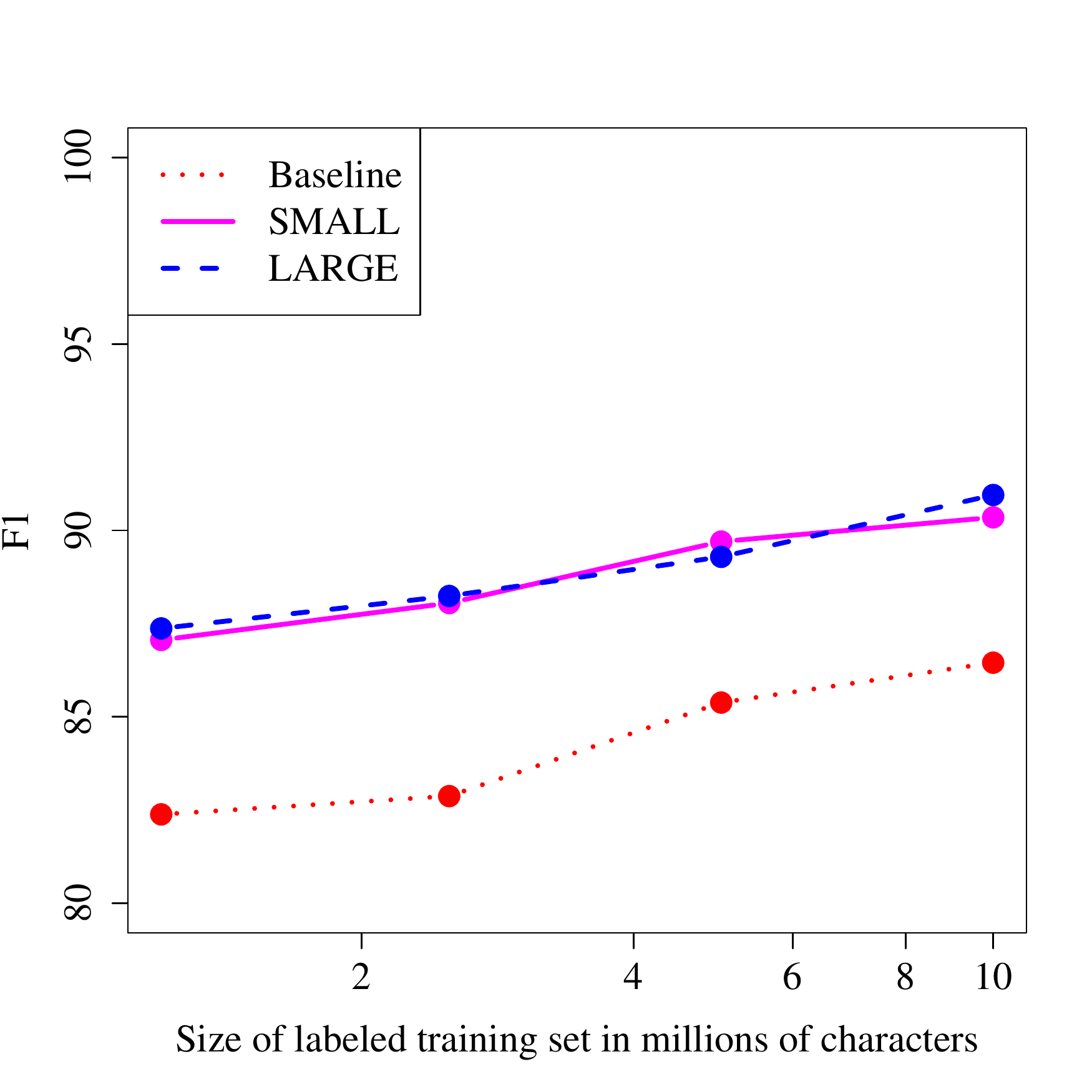}
  \caption{Block segment performance of baseline and augmented feature
    sets on the test set.}
  \label{fig:ablations-test}
\end{figure}

\end{document}